\documentclass[twocolumn]{article}
\usepackage{fullpage}
\usepackage[svgnames,dvipsnames,color,table]{xcolor}
\usepackage{hyperref}
\hypersetup{colorlinks,linkcolor={blue},citecolor={blue},urlcolor={RoyalBlue}} 

\usepackage{amsthm}
\usepackage{amsmath} %
\usepackage{nicefrac}
\usepackage{bm}
\usepackage{booktabs}
\usepackage{amsfonts}
\usepackage{bbm}
\usepackage{mleftright}
\usepackage{dsfont}
\usepackage{amscd,amssymb,amsmath,amsthm,bbold}
\usepackage{graphicx}
\usepackage{stfloats}
\usepackage{multirow}
\usepackage{wrapfig}
\usepackage[numbers]{natbib}

\usepackage{pifont}
\usepackage{microtype}
\usepackage{enumitem}
\usepackage{xfrac}
\usepackage{parskip}
\usepackage{numprint}
\usepackage{booktabs} %

\usepackage{transparent}

\usepackage{algorithm}
\usepackage{algorithmic}
\usepackage{listings}

\usepackage{xspace}
\usepackage{subcaption}
\usepackage{etoolbox}
\usepackage{comment}
\usepackage{etoc}
\usepackage{wrapfig}
\usepackage{placeins}
\usepackage{makecell}
\usepackage{lipsum,xcolor}

\usepackage{microtype}
\usepackage{booktabs} %
\usepackage{makecell}

\usepackage{amsmath}
\usepackage{amssymb}
\usepackage{mathtools}
\usepackage{amsthm}
\usepackage{chngcntr}
\usepackage{xcolor}

\usepackage{nicefrac}
\usepackage{bm}
\usepackage{booktabs}
\usepackage{amsfonts}
\usepackage{bbm}
\usepackage{mleftright}
\usepackage{dsfont}
\usepackage{placeins}
\usepackage{amscd,amssymb,amsmath,amsthm,bbold}
\usepackage[normalem]{ulem}
\usepackage[font=small]{caption}
\usepackage{bbm}

\usepackage{epigraph}

\usepackage{adjustbox}
\usepackage{array}

\newcolumntype{R}[2]{%
    >{\adjustbox{angle=#1,lap=\width-(#2)}\bgroup}%
    l%
    <{\egroup}%
}

\theoremstyle{plain}

\theoremstyle{definition}

\theoremstyle{remark}

\mathchardef\mhyphen="2D

\usepackage[disable,textsize=tiny]{todonotes}

\begin{document}

\date{}
\title{Replacing softmax with ReLU in Vision Transformers}
\author{
Mitchell Wortsman \hspace{.5em} 
Jaehoon Lee \hspace{.5em} 
Justin Gilmer \hspace{.5em} 
Simon Kornblith \\
Google DeepMind
}

\maketitle

\begin{abstract}
Previous research observed accuracy degradation when replacing the attention softmax with a point-wise activation such as ReLU.
In the context of vision transformers, we find that this degradation is mitigated when dividing by sequence length.
Our experiments training small to large vision transformers on ImageNet-21k indicate that ReLU-attention can approach or match the performance of softmax-attention in terms of scaling behavior as a function of compute.
\end{abstract}

\section{Introduction}
\begin{figure*}[b]
    \centering
    \includegraphics[width=0.86\textwidth]{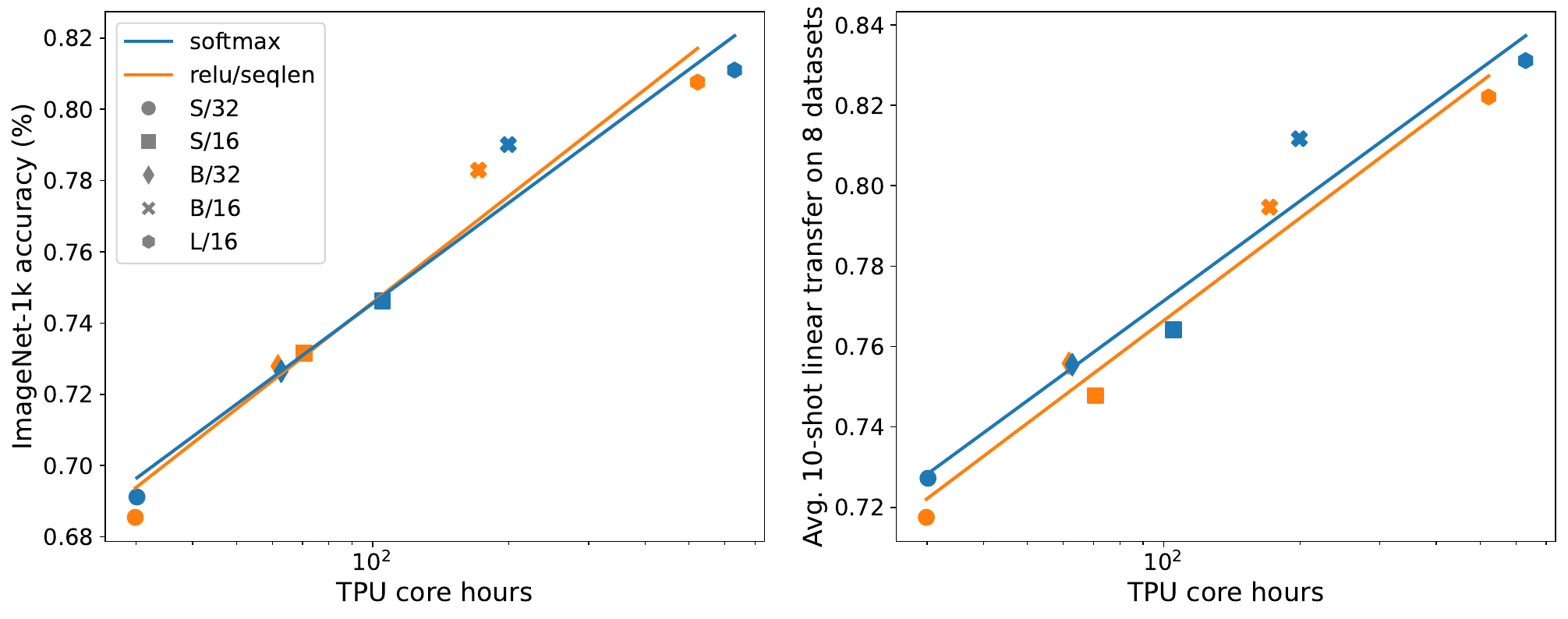}
    \caption{Replacing $\mathsf{softmax}$ with $\mathsf{relu}/\mathsf{seqlen}$ approaches or matches the scaling performance of traditional attention for vision transformers~\cite{dosovitskiy2021an} with qk-layernorm~\cite{dehghani2023scaling}.
    This figure displays results for small to large vision transformers trained on ImageNet-21k~\cite{deng2009imagenet} for 30 epochs. 
    We report ImageNet-1k accuracy for ImageNet-21k models by taking the top class among those that are in ImageNet-1k, without fine-tuning.
    Attention with ReLU can be parallelized over the sequence length dimension with less gather operations than softmax attention.
    }
    \label{fig:fig1}
\end{figure*}

The transformer architecture~\cite{vaswani2017attention} is ubiquitous in modern machine learning. 
Attention, a central component of the transformer~\cite{bahdanau2014neural}, includes a softmax which produces a probability distribution over tokens.
Softmax is costly due to an exponent calculation and a sum over sequence length which makes parallelization challenging~\cite{rabe2021self, dao2022flashattention}.

In this report we explore point-wise alternatives to the softmax operation which do not necessarily output a probability distribution.
As a highlight, we observe that attention with ReLU divided by sequence length can approach or match traditional softmax attention in terms of scaling behavior as a function of compute for vision transformers.
This result presents new opportunities for parallelization, as ReLU-attention can be parallelized over the sequence length dimension with fewer gather operations than traditional attention.

\section{Related work}

Previous research has explored substituting softmax with ReLU~\cite{shen2023study, hron2020infinite} or squared ReLU~\cite{hua2022transformer}.
However, these approaches do not divide by sequence length, which we experimentally find is important to reach accuracy comparable to softmax.
In addition, previous research~\cite{li2022robust} has replaced softmax while still requiring normalization over the sequence length axis to ensure the attention weights sum to one.
This retains the downside of requiring a gather.
After writing an initial version of this note, it was brought to our attention that the variant of ReLU-atttention we study was also explored with a theoretical motivation~\cite{bai2023transformers, fu2023can}.

Moreover, there is extensive literature which removes activation functions altogether so that attention is linear~\cite{pmlr-v119-katharopoulos20a, lu2021soft, koohpayegani2022sima}, which is useful for long sequence lengths.\footnote{Concretely, with linear attention, the order of matrix multiplies can be switched from $(qk^\top)v$ to $q(k^\top v)$ which changes the compute required from $O(dL^2)$ to $O(d^2L)$ where $q, k, v \in \mathbb{R}^{L \times d}$ are the queries, keys, and values and $L$ is sequence length.}
In our experiments, removing the activation entirely reduced accuracy.

\section{Method}
\begin{figure*}
    \centering
    \includegraphics[width=1\textwidth]{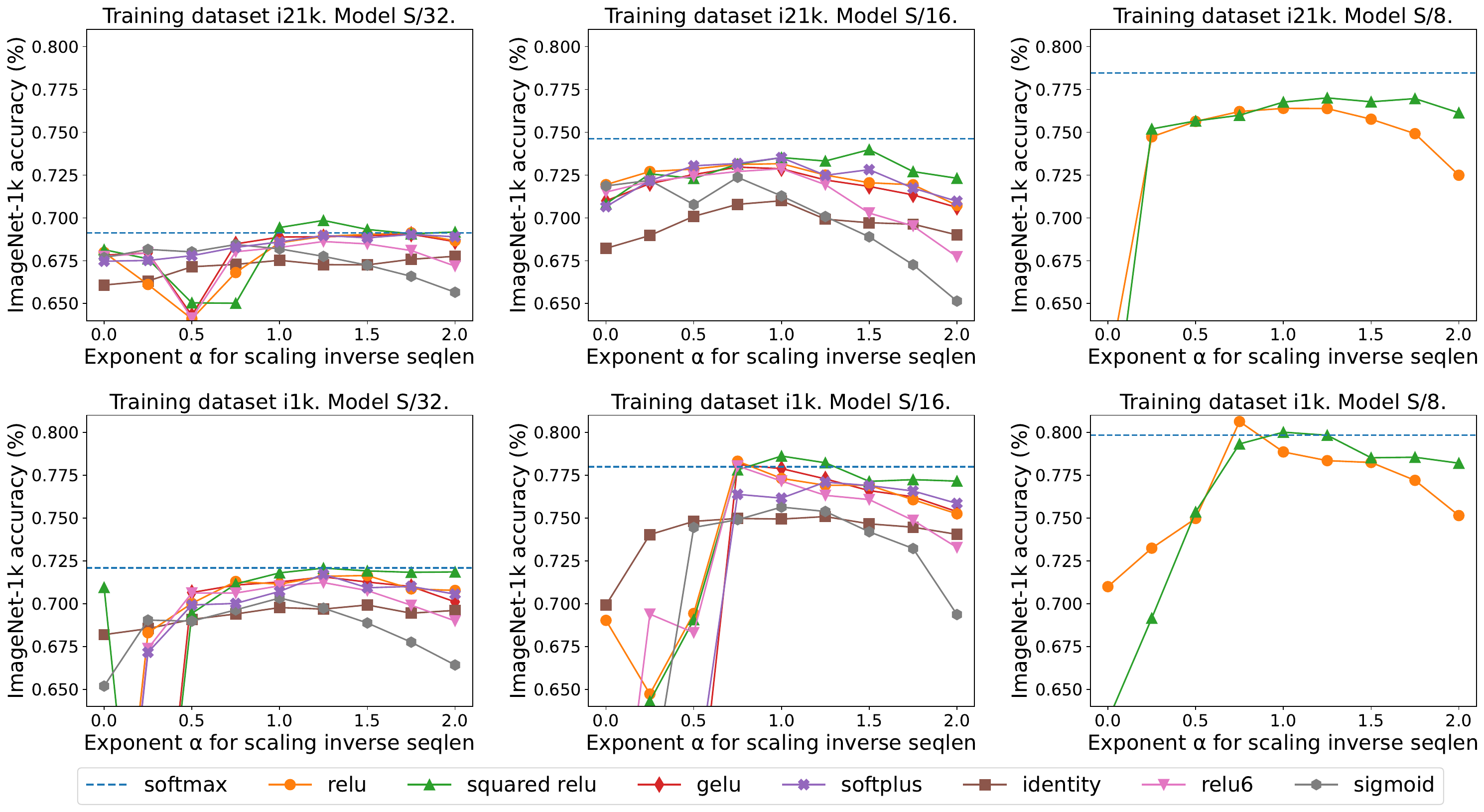}
    \caption{Replacing softmax with $L^{-\alpha} h$ where $h \in \{\mathsf{relu}, \mathsf{relu}^2, \mathsf{gelu}, \mathsf{softplus}, \mathsf{identity}, \mathsf{relu6}, \mathsf{sigmoid}\}$ and $L$ is sequence length.
    We typically observe the best results when $\alpha$ is close to 1.
    There is no clear best non-linearity at $\alpha \approx 1$, so we use ReLU in our main experiment for its speed.}
    \label{fig:fig2}
\end{figure*}

\textbf{Attention.}
Attention transforms $d$-dimensional queries, keys, and values $\{q_i, k_i, v_i\}_{i=1}^L$ with a two step procedure. First, attention weights $\alpha_{ij}$ are produced via
\begin{align}\label{eq:att}
\alpha_{ij} = \phi\left(\frac{1}{\sqrt{d}}\left[q_i^\top k_1, ..., q_i^\top k_L \right]\right)_j,%
\end{align}
where $\phi$ is typically $\mathsf{softmax}$.
Next, the attention weights are used to compute outputs $o_i = \sum_{j = 1}^L \alpha_{ij} v_j$.
This report explores point-wise alternatives to $\phi$. 

\textbf{ReLU-attention.} We observe that $\phi = L^{-1} \mathsf{relu}$ is a promising alternative to $\phi = \mathsf{softmax}$ in Equation~\ref{eq:att}. We refer to attention with $\phi = L^{-1} \mathsf{relu}$ as ReLU-attention.

\textbf{Scaled point-wise attention.} More generally, our experiments will explore $\phi = L^{-\alpha} h$ for $\alpha \in [0, 1]$ and $h \in \{\mathsf{relu}, \mathsf{relu}^2, \mathsf{gelu}, \mathsf{softplus}, \mathsf{identity}, \mathsf{relu6}, \mathsf{sigmoid}\}$ \cite{dahl2013improving, hendrycks2016gaussian}.

\textbf{Sequence length scaling.} We observe that scaling by a term involving sequence length $L$ is beneficial for high accuracy.
This scaling is absent from prior work which removes softmax~\cite{hua2022transformer, koohpayegani2022sima}.
While the central justification for sequence length scaling is empirical, we provide brief analytical motivation.

Transformers are currently designed with softmax attention for which $\sum_{j=1}^L \alpha_{ij} = 1$.
This implies that $\mathds{E}_j[\alpha_{ij}] = L^{-1}$.
While it is unlikely that this is a necessary condition, $\phi = L^{-1}\mathsf{relu}$ does ensure that $\mathds{E}_j[\alpha_{ij}]$ is $O(L^{-1})$ at initialization.
Preserving this condition may alleviate the need to change other hyperparameters when replacing softmax.

At initialization the elements of $q$ and $k$ are $O(1)$ and so $\frac{\langle q_i, k_j\rangle}{\sqrt{d}}$ will also be $O(1)$.
Activation functions such as ReLU preserve $O(1),$\footnote{With the exception of squared ReLU.} and so a factor $L^{-1}$ is necessary for  $\mathds{E}_j[\alpha_{ij}]$ to be $O(L^{-1})$.

\section{Experiments}

\begin{figure*}[t]
    \centering
    \includegraphics[width=1\textwidth]{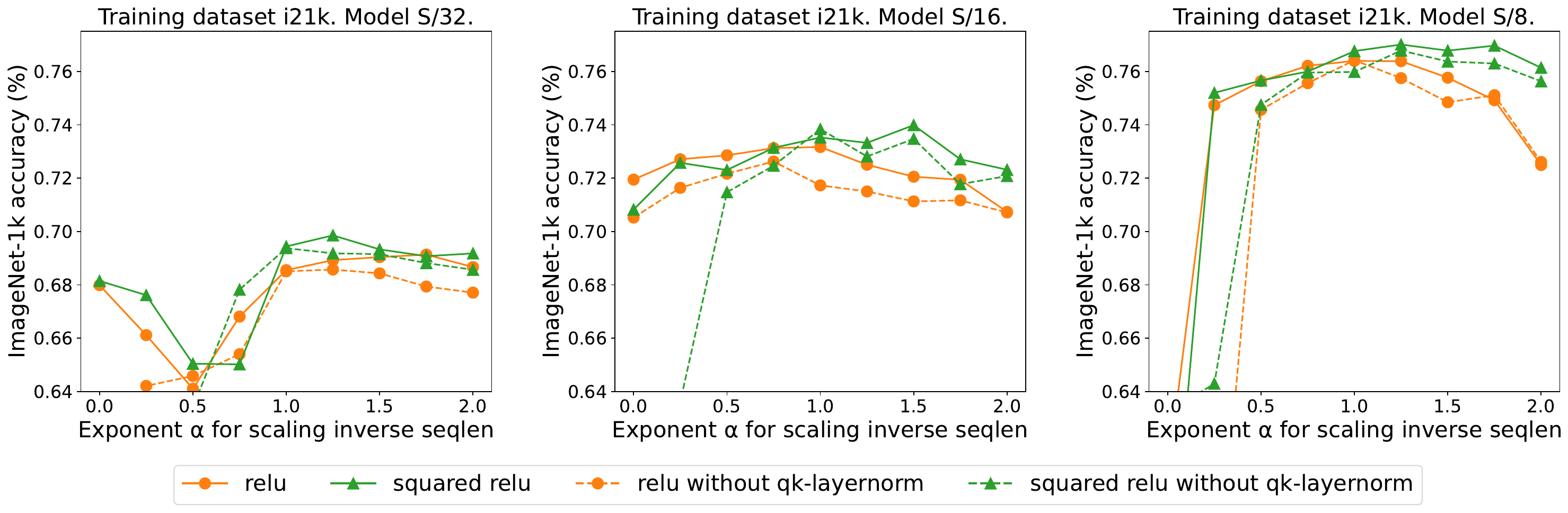}
    \caption{The effect of removing qk-layernorm~\cite{dehghani2023scaling} on attention with ReLU and squared ReLU scaled by $L^{-\alpha}$ where $L$ is sequence length.
    Results are shown for the S/32, S/16, and S/8 vision transformer models~\cite{dosovitskiy2021an,vit_baseline} trained on ImageNet-21k.}
    \label{fig:fig3}
\end{figure*}

\begin{figure*}[t]
    \centering
    \includegraphics[width=1\textwidth]{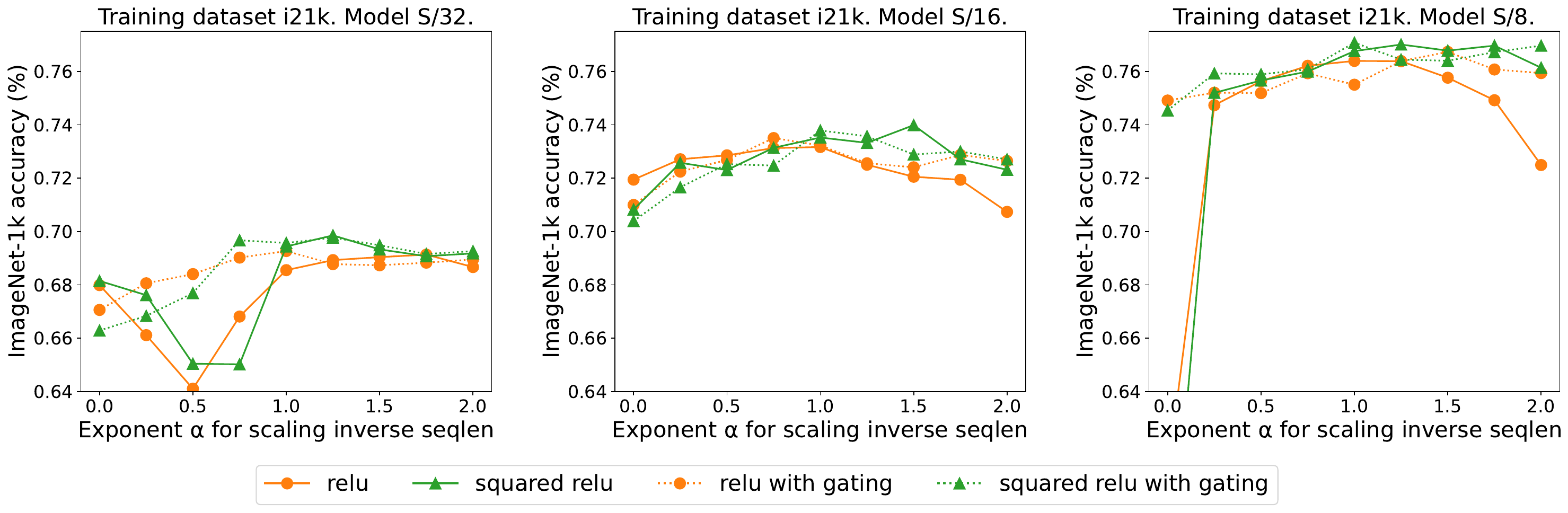}
    \caption{The effect of using a gated attention unit~\cite{hua2022transformer} on attention with ReLU and squared ReLU scaled by $L^{-\alpha}$ where $L$ is sequence length.
    Results are shown for the S/32, S/16, and S/8 vision transformer models~\cite{dosovitskiy2021an,vit_baseline} trained on ImageNet-21k.}
    \label{fig:fig4}
\end{figure*}

\textbf{Experimental setup.} Our experiments use ImageNet-21k and ImageNet-1k~\cite{deng2009imagenet} training configurations from the BigVision codebase~\cite{vit_baseline} without modifying hyperparameters.\footnote{For ImageNet1k we use the base config \url{https://github.com/google-research/big_vision/blob/main/big_vision/configs/vit_i1k.py}. For ImageNet21k we use the base config \url{https://github.com/google-research/big_vision/blob/main/big_vision/configs/vit_i21k.py}.}
In our experiments on ImageNet-21k we train for 30 epochs, and in our experiments on ImageNet-1k we train for 300 epochs.
As a result, both training runs use a roughly similar number of steps of around 9e5.
We use ViTs with qk-layernorm~\cite{dehghani2023scaling} as this was previously observed to be necessary to prevent instability when scaling model size. However, we ablate that this is not an important component at the scales we test.
We use i21k and i1k to mean ImageNet-21k and ImageNet-1k, respectively, and report ImageNet-1k accuracy for ImageNet-21k models by taking the top class among those that are in ImageNet-1k, without fine-tuning.
When evaluating transfer performance on downstream tasks we use a 10-shot linear probe averaged over three seeds.
The downstream tasks are Caltech Birds~\cite{welinder2010caltech}, Caltech-101~\cite{fei2004learning}, Stanford Cars~\cite{cars}, CIFAR-100~\cite{krizhevsky2009learning}, DTD~\cite{dtd}, ColHsit~\cite{kather2016collection}, Pets~\cite{parkhi2012cats}, and UC Merced~\cite{yang2010bag}.

\textbf{Main experiment.} Figure~\ref{fig:fig1} illustrates that ReLU-attention matches the scaling trends for softmax attention for ImageNet-21k training.
On the $x$-axis we display the total core hours required for the experiment.
As an advantage, ReLU-attention enables parallelization over the sequence length dimension with fewer gather operations than softmax attention.

\textbf{Effect of sequence length scaling.} Figure~\ref{fig:fig2} examines the effect of sequence length scaling for various point-wise alternatives to softmax.
Concretely, we replace softmax with $L^{-\alpha} h$ for $\alpha \in [0, 1]$ and $h \in \{\mathsf{relu}, \mathsf{relu}^2, \mathsf{gelu}, \mathsf{softplus}, \mathsf{identity}\}$.
On the $x$-axis we display $\alpha$.
The $y$-axis displays accuracy for the S/32, S/16, and S/8 vision transformer models~\cite{dosovitskiy2021an, vit_baseline}.
The best results are typically achieved when $\alpha$ is close to 1.
Since there is not clear best non-linearity, we use ReLU in our main experiment as it is faster.

\textbf{Effect of qk-layernorm.} Our main experiments use qk-layernorm~\cite{dehghani2023scaling} in which queries and keys are passed through LayerNorm~\cite{ba2016layer} before computing attention weights.
We use qk-layernorm by default as it was found to be necessary to prevent instability when scaling up model size~\cite{dehghani2023scaling}.
Figure~\ref{fig:fig3} shows the effect of removing qk-layernorm.
The results indicate that qk-layernorm does not have a large effect for these models, but this may change at scale.

\textbf{Effect of adding a gate.} 
Previous work removing softmax adds a gated unit and does not scale by sequence length~\cite{hua2022transformer}.
Concretely, in the gated attention unit~\cite{hua2022transformer} an extra projection produces output which is combined through elementwise-multiplication before the out projection.
In Figure~\ref{fig:fig4} we investigate whether the presence of a gate removes the need for sequence length scaling.
Overall we observe that the best accuracy is still achieved with sequence length scaling, with or without the gate.
Note that gating increases the core hours required for the experiment by roughly 9.3\% for the S/8 model with ReLU.

\section{Conclusion}

This report leaves many open questions.
In particular, we are unsure why the factor $L^{-1}$ improves performance or if this term could be learned.
Moreover, it is likely that there is a better activation function that we do not explore.

\subsubsection*{Acknowledgements}

We thank Lucas Beyer, Mostafa Dehghani, and David Fleet for their helpful comments and suggestions.

We thank the members of the Google DeepMind PAGI team for their support of this effort, 
Jascha Sohl-dickstein, Noah Fiedel, Aaron Parisi, Abhishek Kumar, Alex Alemi, Alex Rizkowsky,  Avi Singh, Azade Nova, Ben Adlam, Bernd Bohnet, Daniel Freeman,  Gamaleldin Elsayed, Gaurav Mishra, Hanie Sedghi, Isabelle Simpson, Izzeddin Gur, JD Co-Reyes, James Harrison, Jeffrey Pennington, Jiri Hron, Kathleen Kenealy, Kelvin Xu, Kevin Swersky, Kshiteej Mahajan, Laura Culp, Lechao Xiao, Max Bileschi, Merrie Morris, Roman Novak, Rosanne Liu, Sharad Vikram, Tris Warkentin, Yundi Qian.
\FloatBarrier

{\small
\bibliographystyle{plainnat}
\bibliography{main}
}
\clearpage
\appendix

\end{document}